\documentclass[letterpaper]{article} 
\usepackage{aaai2026}  
\usepackage{times}  
\usepackage{helvet}  
\usepackage{courier}  
\usepackage[hyphens]{url}  
\usepackage{graphicx} 
\usepackage{natbib}  
\usepackage{caption} 
\usepackage{multirow}
\usepackage{makecell}
\usepackage{color}
\usepackage{amsmath}
\usepackage{amssymb}
\usepackage{algorithm}
\usepackage{algorithmic}

\usepackage{newfloat}
\usepackage{listings}
\DeclareCaptionStyle{ruled}{labelfont=normalfont,labelsep=colon,strut=off} 
\floatstyle{ruled}
\newfloat{listing}{tb}{lst}{}
\floatname{listing}{Listing}
\title{Exploring Task-Solving Paradigm for Generalized Cross-Domain Face Anti-Spoofing via Reinforcement Fine-Tuning}
\author{
    Fangling Jiang \textsuperscript{\rm 1} \equalcontrib ,
    Qi Li \textsuperscript{\rm 2,3} \equalcontrib,
    Weining Wang \textsuperscript{\rm 4} ,
    Gang Wang\textsuperscript{\rm 4} ,
    Bing Liu \textsuperscript{\rm 1} ,
    Zhenan Sun \textsuperscript{\rm 2,3}
}
\affiliations{
	\textsuperscript{\rm 1} School of Computer Science, University of South China, Hengyang, China. jfl@usc.edu.cn\\
	\textsuperscript{\rm 2} New Laboratory of Pattern Recognition, MAIS, CASIA, Beijing, China\\
	\textsuperscript{\rm 3} School of Artificial Intelligence, UCAS, Beijing, China\\	
	\textsuperscript{\rm 4} The Laboratory of Cognition and Decision Intelligence for Complex Systems, CASIA, Beijing, China\\

}

\usepackage{bibentry}

\begin{document}

\maketitle

\begin{abstract}
Recently the emergence of novel presentation attacks has drawn increasing attention to face anti-spoofing. However, existing methods tend to memorize data patterns from the training set, resulting in poor generalization to unknown attack types across different scenarios and limited interpretability. 
To address these challenges, this paper presents a reinforcement fine-tuning-based face anti-spoofing method that stimulates the capabilities of multimodal large language models to think and learn how to solve the anti-spoofing task itself, rather than relying on the memorization of authenticity patterns. 
We design verifiable class consistent reward and reasoning consistent reward, and employ a GRPO-based optimization strategy to guide the model in exploring reasoning policies from multiple perspectives to maximize expected rewards. 
As a result, through iterative trial-and-error learning while retaining only high-reward trajectories, the model distills highly generalizable decision-making rules from the extensive solution space to effectively address cross-domain face anti-spoofing tasks.
Extensive experimental results demonstrate that our method achieves state-of-the-art cross-domain generalization performance. 
It generalizes well to diverse unknown attack types in unseen target domains while providing interpretable reasoning for its authenticity decisions without requiring labor-intensive textual annotations for training.
\end{abstract}


\section{Introduction}
Face anti-spoofing aims to distinguish between real faces and spoof faces presented to a camera, thereby preventing spoof faces from impersonating legitimate users and bypassing face recognition systems. With advancements in fabrication techniques in recent years, a wide array of spoof faces, such as printed photos, replayed videos, masks, and makeup, have emerged in rapid succession. Consequently, face anti-spoofing has garnered significant attention from both industry and academia, particularly in the context of cross-domain face anti-spoofing, which is urgently needed in real-world applications.

\begin{figure}[htb]
	\centering
	\includegraphics[width=0.48\textwidth]{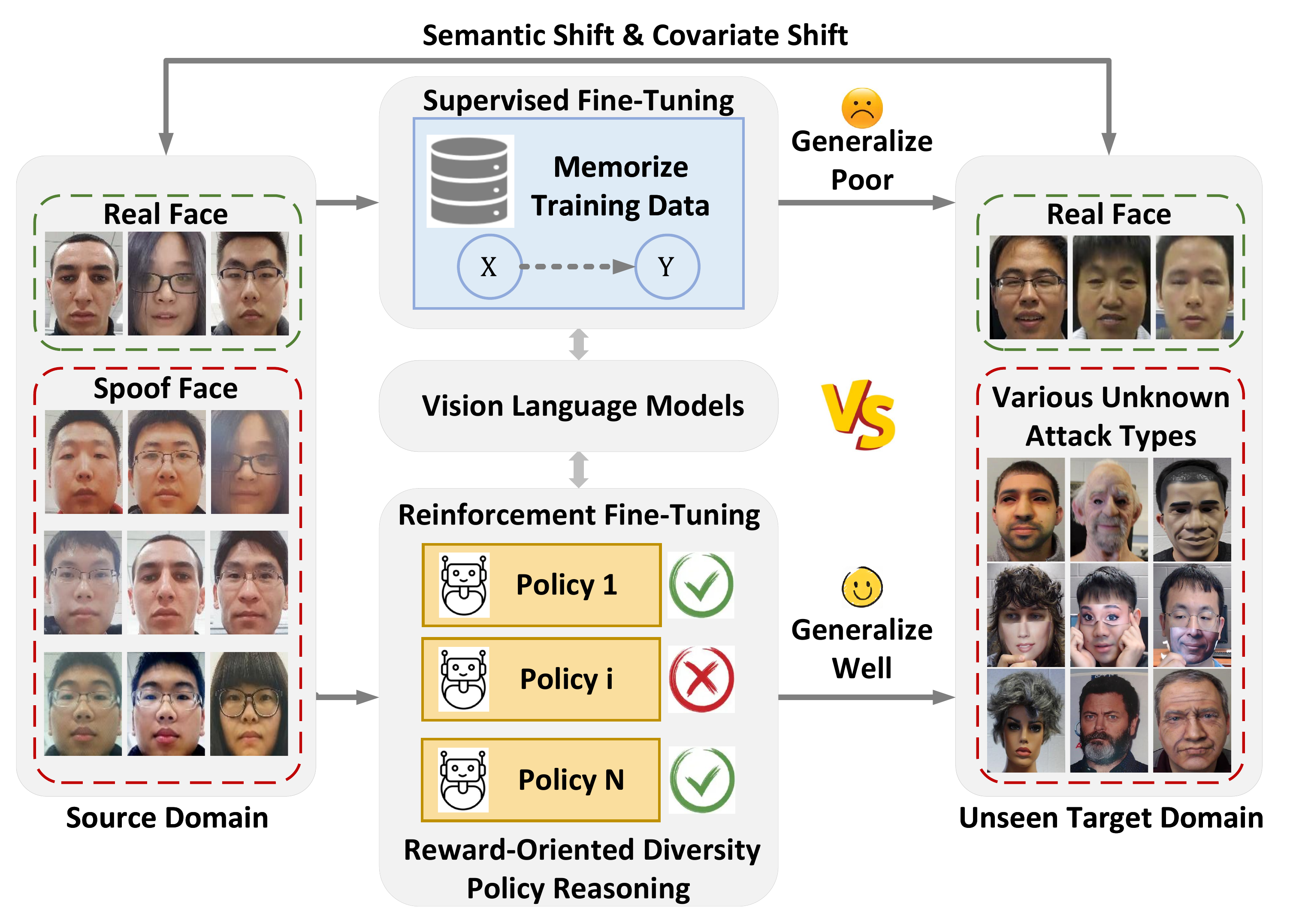}
	\caption{
	In contrast to supervised fine-tuning, which relies on explicit supervision through annotated answers and tends to encourage the model to memorize spoof patterns in the training data to reproduce superficial label forms, our approach guides the vision-language model through reward-based trial-and-error learning. This facilitates autonomous exploration of diverse solution pathways, ultimately enhancing the reasoning and generalization capabilities of the model by enabling it to acquire transferable decision-making policies.
}
\label{fig6}
\end{figure}

In cross-domain face anti-spoofing, the testing data (target domain) is typically unknown and exhibits substantial distribution shifts from the training data (source domain). Such distribution shifts can stem from covariate shifts induced by spoof-irrelevant external factors, such as background, lighting conditions, and recording devices, or from semantic shifts caused by spoof-relevant intrinsic factors, including variations in structure, material, or texture associated with previously unseen attack types in the target domain~\citep{yu2021deep,jiang2024open}.

Traditional methods often generalize poorly under covariate and semantic shifts in cross-domain scenarios~\cite{dharmawan2024towards}. To address this issue, numerous studies have introduced domain generalization techniques into face anti-spoofing, typically assuming that the target domain shares the same attack types as the source domain. These approaches focus on enhancing generalization to covariate shifts through strategies such as domain alignment~\citep{jia2020single}, feature disentanglement~\citep{yang2024generalized}, and meta-learning~\citep{jia2021dual}. However, given the unpredictability of attack types in real-world target domains, where both covariate and semantic shifts are likely to co-exist, recent works have proposed open-set augmentation at the image and embedding levels~\citep{jiang2024open,ge2024difffas}, as well as one-class anomaly detection frameworks~\citep{huang2024one}, to concurrently address both types of shifts. Vision-language models~\cite{zhang2024vision} trained on large-scale data encapsulate extensive general knowledge. Recently, numerous studies have successfully adapted these models to the face anti-spoofing task through prompt learning and supervised fine-tuning~\citep{liu2024cfpl,srivatsan2023flip,ozgur2025foundpad}, significantly improving their ability to generalize to unseen target domains.

Nevertheless, supervised fine-tuning requires the labor-intensive and time-consuming annotation of rich, explicit textual answers. Moreover, such explicit supervision often leads the model to memorize authenticity patterns present in the training data in order to reproduce the surface form of the labels~\citep{chu2025sft}. This tendency increases the risk of the model exploiting domain-specific features, thereby compromising its ability to generalize under significant covariate and semantic shifts in unseen target domains.

Inspired by the educational philosophy of \textit{teaching one to fish rather than giving one a fish}, as illustrated in \figurename~\ref{fig6}, this study employs reinforcement learning to guide vision-language models in acquiring an intrinsic classification mechanism for discerning real and spoof faces, rather than relying on pattern memorization and answer imitation. Specifically, we design class consistent reward and reasoning consistent reward tailored to the face anti-spoofing task.
Through Group Relative Policy Optimization~\citep{shao2024deepseekmath} (GRPO)-based iterative optimization strategy, the model is encouraged to explore diverse reasoning policies from multiple perspectives to maximize expected rewards. 
This process drives the model to extract the most task-relevant and discriminative information from images while ignoring irrelevant details, and to optimize its behavior directly toward reward objectives rather than the superficial form of annotated answers. By exploring various policies and retaining only those that yield high rewards, the model effectively distills robust decision-making rules from a vast solution space. 
These rules exhibit strong generalizability for cross-domain face anti-spoofing and enable better adaptation to significant covariate and semantic shifts in unseen target domains.

The main contributions of this paper are summarized as follows:
\begin{enumerate}    
	\item We propose a reinforcement fine-tuning-based face anti-spoofing method that learns transferable task-solving logic, achieving strong generalization to significant covariate and semantic shifts in cross-domain unseen target domains.
	\item We use only real or spoof labels, eliminating the need to construct large-scale textual reasoning annotations, while enabling interpretable decision-making reasoning for real and spoof face classification.
	\item Extensive experiments demonstrate that our approach achieves state-of-the-art performance for cross-domain face anti-spoofing. 
	It effectively defends against diverse unknown attack types such as makeup and masks made from different materials in unseen target domains.
\end{enumerate}

\begin{figure*}[t]
	\centering
	\includegraphics[width=0.95\textwidth]{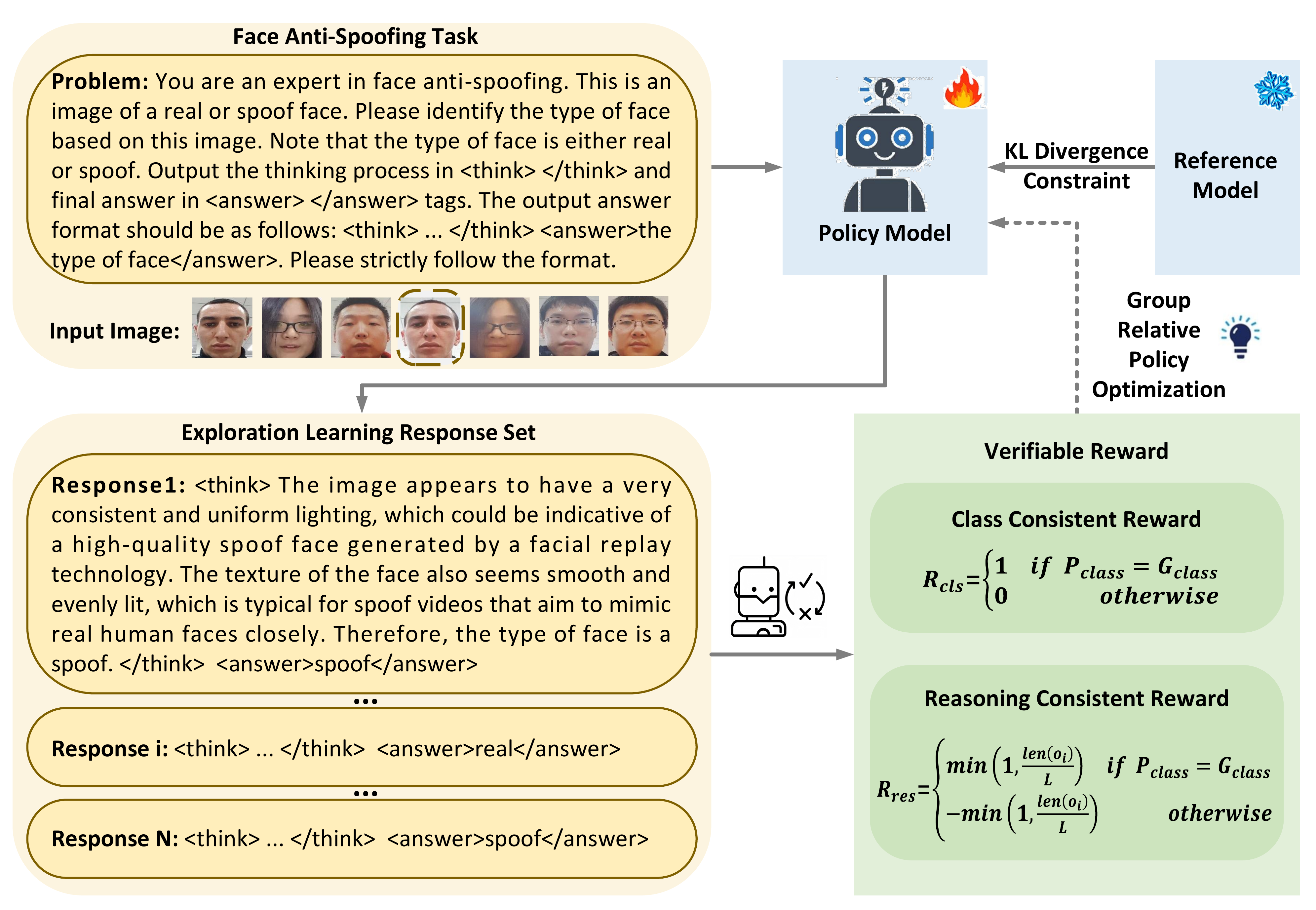} 
	\caption{Overview of the reinforcement fine-tuning framework for generalized cross-scenario face anti-spoofing. The framework introduces class consistent reward and reasoning consistent reward to guide the model toward accurate category predictions while maintaining reasonable reasoning length. A GRPO-based policy optimization mechanism is employed to encourage the model to explore diverse reasoning policies from multiple perspectives in order to maximize expected rewards. Through this process, the model distills robust decision-making rules from a vast solution space, leading to strong generalization across significant covariate and semantic shifts in unseen target domains.
}
	\label{fig7}
\end{figure*}
\section{Related Work}\label{sec2}
\subsection{Cross-Domain Face Anti-Spoofing}
Common cross-domain face anti-spoofing methods include domain adaptation, domain generalization, and one-class anomaly learning-based methods. Early domain adaptation methods~\citep{jia2021unified} focused on aligning the feature distributions between labeled source domains and unlabeled target domains. Given the difficulty of obtaining source domain data in many real-world scenarios, source-free~\citep{liu2022source,li2025optimal} and test-time adaptation~\citep{huang2023test} approaches have emerged. However, accessing even unlabeled target domain data is often challenging. As a result, many studies have adopted the domain generalization paradigm, which does not require target domain data and has been explored through various approaches, including domain alignment\citep{li2018learning,shao2019multi,jia2020single,wang2024csdg,kong2024dual,le2024gradient,liu2025dual,hu2024rethinking}, meta-learning~\citep{jia2021dual,zhang2024domain} , disentangled representation learning\citep{wang2022domain,yang2024generalized,ma2024dual}, prompt learning~\citep{srivatsan2023flip,hu2024fine,liu2024cfpl,wang2024tf,guo2024style,liu2024bottom,fang2024vl,guo2025domain}, multimodal learning~\citep{lin2025reliable,chen2025mmfas} and data augmentation~\citep{cai2022learning,cai2024towards,ge2024difffas}.

Domain generalization typically assumes the availability of multiple source domains and that attack types are consistent between source and target domains. To address these limitations, some works have proposed open-set domain generalization approaches~\citep{jiang2024open,dong2021open} that aim to build models capable of generalizing to unknown attack types using only a limited number of source domains. Considering the high diversity of spoof faces and the difficulty of collecting a comprehensive set of spoofed samples for training, some studies have framed face anti-spoofing as an anomaly detection problem~\citep{huang2024one,narayan2024hyp,huang2025slip}. These methods focus on learning from one-class real face data to build models that can generalize to a wide range of unknown spoof attacks.

\subsection{MultiModal Large Models in Face Anti-Spoofing}
Many face anti-spoofing studies leverage the general capabilities of multimodal large models to enhance generalization and interpretability. Common approaches include fine-grained learnable prompt tuning~\citep{hu2024fine,liu2024cfpl,mu2023teg,wang2024tf,guo2024style,liu2024bottom,fang2024vl,guo2025domain}, fixed prompt combined with supervised fine-tuning~\citep{srivatsan2023flip}, and parameter-efficient supervised fine-tuning~\citep{ozgur2025foundpad}. Some works~\citep{zhang2025interpretable,wang2025faceshield} utilize large language models with human-verified generated textual explanations as training data to fine-tune multimodal large language models for interpretable decision-making in face anti-spoofing.

In contrast to previous methods, we employ reinforcement fine-tuning to uncover the general capabilities of multimodal large language models. This approach eliminates the need for labor-intensive textual annotations, enabling the model to actively explore solutions to face anti-spoofing tasks while simultaneously generating interpretable reasoning for its decisions.

\section{Proposed Method}
\label{sec3}
\subsection{Problem Definition}
Given a source domain $\mathcal{D}^s=\left\{(x_{i},y_{i})\right\}_{i=1}^{N}$, where $x_i$ denotes the $i$-th training sample and $y_i$ represents its corresponding class label, with $y \in \mathcal{C}$ (the label space), our objective is to train a face anti-spoofing model based on $\mathcal{D}^s$ that generalizes effectively to unseen target domains $\mathcal{D}^{t}=\left\{x_{i}^{t}\right\}_{i=1}^{N^{m}}$. These target domains exhibit significant covariate and semantic shifts relative to the source domain. Furthermore, the face anti-spoofing model is expected to provide interpretable reasoning behind its authenticity decisions.

Unlike conventional training paradigms that encourage memorization of mappings between image patterns and labels in the training data, we advocate for learning problem-solving strategies tailored to face anti-spoofing by a reinforcement fine-tuning framework, as illustrated in \figurename~\ref{fig7}. The model is guided to acquire policy-level knowledge that enables it to adaptively generalize to novel data patterns and attack types in unseen domains.

\subsection{Preliminary of Group Relative Policy Optimization}
Group Relative Policy Optimization (GRPO) is a widely adopted policy optimization algorithm in reinforcement learning, distinguished by its core principle of refining the learning process through comparative evaluation of relative values among strategies within the same group, rather than relying on conventional critic models to assess the absolute value of individual policies.  
Specifically, given a problem $q$, the old policy model $\pi_{\theta_{\text{old}}}$ initially samples multiple candidate policies to form a policy group $\{o_1,o_2,...,o_N\}$. These policies are then evaluated by rule-based reward functions, generating \( N \) reward scores $\{r_1,r_2,...,r_N\}$. Subsequently, these rewards are normalized by subtracting the group mean $mean({r_1,...,r_N})$ and dividing by the group standard deviation $std({r_1,...,r_N})$.
The resulting normalized rewards serve as relative advantages, which are used to update the policy model $\pi_{\theta}$ by maximizing the following objective function: 
\begin{equation}\label{equ2}
	\begin{aligned}
		\mathcal{J}_{\text{GRPO}}(\theta) = &\mathbb{E}_{[q \sim Q, \{o_i\}_{i=1}^N \sim \pi_{\theta_{\text{old}}}(o|q)]} \frac{1}{N} \sum_{i=1}^N \frac{1}{|o_i|} \sum_{t=1}^{|o_i|} \\
		&\big\{ \min \left[ \frac{\pi_{\theta}^{i, t}}{\pi_{\theta_{\text{old}}}^{i, t}}A_{i, t}, \text{clip} \left( \frac{\pi_{\theta}^{i, t}}{\pi_{\theta_{\text{old}}}^{i, t}}, 1 - \epsilon, 1 + \epsilon \right)A_{i, t} \right] \\
		&-\beta \cdot \mathbb{D}_{\text{KL}} \left[ \pi_{\theta} \| \pi_{\text{ref}} \right]\big\},
	\end{aligned}
\end{equation}
where $\epsilon$ and $\beta$ are hyper-parameters, the advantage $A_{i, t}$ is defined as
\begin{equation}\label{equ1}
	\begin{aligned}
		A_{i, t} =  \frac{r_i-mean({r_1,...,r_N})}{std({r_1,...,r_N})}. 
	\end{aligned}
\end{equation}
Additionally, to regulate the magnitude of policy updates, a Kullback-Leibler (KL) divergence constraint $\mathbb{D}_{\text{KL}}$ is incorporated, ensuring the updated policy model $\pi_{\theta}$ does not deviate excessively from the reference policy model $\pi_{\text{ref}}$, thereby mitigating the risk of policy collapse.

\subsection{Verifiable Rewards for Face Anti-Spoofing}
To learn decision policies with strong generalization capabilities toward unseen target domains by GRPO, it is essential to design some verifiable reward functions tailored to the face anti-spoofing task. We employ three different rewards, namely class consistent reward, reasoning consistent reward, and format reward.

\textbf{Format Reward.}
To ensure the model not only classifies facial authenticity but also provides interpretable reasoning for its decisions, we mandate that its response comprises two distinct components: the reasoning process enclosed within $<thinking>...</thinking>$ tags and the facial classification outcome enclosed within $<answer>...</answer>$ tags.  
A format reward is introduced to quantitatively assess the model’s adherence to this prescribed output structure:
\begin{equation}\label{equ3}
	\begin{aligned}
		R_{\text{format}} = \begin{cases}
			1, & \text{if response matches format}, \\
			0, & \text{if response does not match format}.
		\end{cases}
	\end{aligned}
\end{equation}

\textbf{Class Consistent Reward.} The class consistency reward ensures that the policy maintains discriminative feature representations aligned with real and spoof classes. We extract the predicted face class $P_{class}$ enclosed within $<answer>...</answer>$ and compare it with the ground truth class $G_{class}$. If the predicted face class matches the ground truth class, a reward is granted; otherwise, no reward is given. Since the response of models is in textual form, we define the ground truth classes as $real$ and $fake$. The formalized reward calculation is as follows:
\begin{equation}\label{equ4}
	\begin{aligned}
		R_{\text{cls}} = \begin{cases}
			1, & \text{if    } P_{class}=G_{class}, \\
			0, & \text{if    } P_{class} \neq G_{class}.
		\end{cases}
	\end{aligned}
\end{equation}

\textbf{Reasoning Consistent Reward.} The reasoning consistent reward guides the model to achieve a balanced relationship between reasoning and accurate category prediction. When the model correctly predicts the face class, its reasoning is likely to support the correct decision. In such cases, we encourage the model to produce as detailed reasoning as possible by applying a positive reward that increases with the length of the reasoning. Conversely, when the predicted category is incorrect, excessive reasoning may reinforce the error and mislead the model further. Therefore, we apply a penalizing reward to discourage overly long reasoning in incorrect predictions, guiding the model to generate concise reasoning instead. The specific calculation of the reasoning consistent reward is as follows:
\begin{equation}\label{equ5}
	\begin{aligned}
		R_{\text{res}} = \begin{cases}
			min(1,\frac{len(o_i)}{L}), & \text{if    } P_{class}=G_{class}, \\
			-min(1,\frac{len(o_i)}{L}), & \text{if    } P_{class} \neq G_{class},
		\end{cases}
	\end{aligned}
\end{equation}
where $L$ denotes the expected maximum length, and $\text{len}(\cdot)$ represents the length computation function.

Together, the three rewards are combined to $R_{\text{all}}$ as defined in
\begin{equation}\label{equ6}
	\begin{aligned}
		R_{\text{all}} = R_{\text{format}}+R_{\text{cls}}+R_{\text{res}}
	\end{aligned}
\end{equation}
to guide the optimization process toward a generalized and reliable anti-spoofing policy.

\begin{table*}[!tbp]
	\footnotesize
	\centering 
	\begin{tabular}{|p{2.5cm}|p{0.8cm}<{\centering}|p{0.8cm}<{\centering}|p{0.8cm}<{\centering}|p{0.8cm}<{\centering}|p{0.8cm}<{\centering}|p{0.8cm}<{\centering}|p{0.8cm}<{\centering}|p{0.8cm}<{\centering}|p{0.8cm}<{\centering}|p{0.8cm}<{\centering}|p{0.8cm}<{\centering}|p{0.8cm}<{\centering}|
		}
		\hline
		\multirow{2}{*}{Method}&\multicolumn{3}{c|}{CeFa to HQ-WMCA(\%)$\downarrow$}&\multicolumn{3}{c|}{CeFa to SiW-Mv2(\%)$\downarrow$}&\multicolumn{3}{c|}{SURF to HQ-WMCA(\%)$\downarrow$}&\multicolumn{3}{c|}{SURF to SiW-Mv2(\%)$\downarrow$}\\ \cline{2-13}
		& FRR&FAR &HTER&FRR &FAR &HTER& FRR&FAR &HTER&FRR &FAR &HTER   \\ \hline
		MS-LBP&100.00&0.22&50.11&99.48&0.55&50.02&93.35&13.96&53.65&11.60&85.57&48.58\\ \hline
		Color texture&100.00&0.11&50.05&99.87&0.11&49.99&0.00&100.00&50.00&68.17&35.85&52.01\\ \hline
		CNN&35.55&59.54&47.55&75.77&21.64&48.71&100.00&0.00&50.00&93.94&8.55&51.24 \\ \hline
		Flip&19.52&19.13&\underline{19.32}&22.62&22.68&\underline{22.65}&20.12&20.17&\underline{20.14}&15.52&15.54&15.53\\ \hline
		FoundPAD ViT-FS&47.61&47.71&47.66&25.10&25.14&25.12&46.15&46.13&46.14&18.85&19.02&18.93\\ \hline
		FoundPAD FE&49.48&49.56&49.52&30.19&30.16&30.18&46.78&46.62&46.70&20.89&20.87&20.88\\ \hline
		FoundPAD&47.82&47.82&47.82&29.81&29.95&29.88&46.36&46.40&46.38&13.50&13.66&\underline{13.58}\\ \hline
		Ours&7.90&23.61&\textbf{15.75}&4.33&13.44&\textbf{8.89}&6.03&26.94&\textbf{16.48}&0.89&18.58&\textbf{9.74}\\ \hline
	\end{tabular}
	\caption{Cross-domain evaluation results under the four protocols: CeFa to HQ-WMCA, CeFa to SiW-Mv2, CASIA-SURF to HQ-WMCA, and CASIA-SURF to SiW-Mv2.}
	\label{tabcross}
\end{table*}
\begin{table*}[!tbp]
	\footnotesize
	\centering 
	\begin{tabular}{|p{3.9cm}|p{0.8cm}<{\centering}|p{0.8cm}<{\centering}|p{0.8cm}<{\centering}|p{0.8cm}<{\centering}|p{0.8cm}<{\centering}|p{0.8cm}<{\centering}|p{0.8cm}<{\centering}|p{0.8cm}<{\centering}|p{0.8cm}<{\centering}|p{0.8cm}<{\centering}|p{0.8cm}<{\centering}|p{0.8cm}<{\centering}|
		}
		\hline
		\multirow{2}{*}{Method}&\multicolumn{3}{c|}{CeFa to HQ-WMCA(\%)$\downarrow$}&\multicolumn{3}{c|}{CeFa to SiW-Mv2(\%)$\downarrow$}&\multicolumn{3}{c|}{SURF to HQ-WMCA(\%)$\downarrow$}&\multicolumn{3}{c|}{SURF to SiW-Mv2(\%)$\downarrow$}\\ \cline{2-13}
		& FRR&FAR &HTER&FRR &FAR &HTER& FRR&FAR &HTER&FRR &FAR &HTER   \\ \hline
		Qwen2.5-VL-7B-Instruct&0.00&55.34&\underline{27.67}&0.13&59.67&29.90&0.00&55.34&27.67&0.13&59.67&29.90\\ \hline
		Qwen2.5-VL-7B-Instruct SFT&33.68&27.54&30.61&6.30&13.22&\underline{9.76}&4.16&46.73&\underline{25.44}&1.53&56.18&\underline{28.85}\\ \hline
		Ours&7.90&23.61&\textbf{15.75}&4.33&13.44&\textbf{8.89}&6.03&26.94&\textbf{16.48}&0.89&18.58&\textbf{9.74}\\ \hline
	\end{tabular}
	\caption{Comparison of supervised fine-tuning and reinforcement fine-tuning under the four protocols. }
	\label{tabsft}
\end{table*}

\subsection{Training and Inference Process}
We convert the original face anti-spoofing dataset into instruction-style triplets consisting of an image, a question, and an answer for training. The Qwen2.5-VL-7B-Instruct model is selected as the base model for reinforcement fine-tuning, owing to its strong multimodal capabilities, low adaptation threshold for fine-tuning, and robust open ecosystem. The task prompt used during training and inference is illustrated in \figurename~\ref{fig7}. 
During the inference stage, we evaluate the model’s performance by extracting the predicted face class enclosed within $<answer>...</answer>$ from the generated response. In this study, we do not adopt traditional threshold-based evaluation methods; instead, we directly assess the correctness of the prediction by comparing whether the predicted class is equal to the ground truth class.

\section{Experiments}
\label{sec4}
\subsection{Experimental Setups}
\label{sec41}
\begin{figure}[t]
	\centering
	\includegraphics[width=0.5\textwidth]{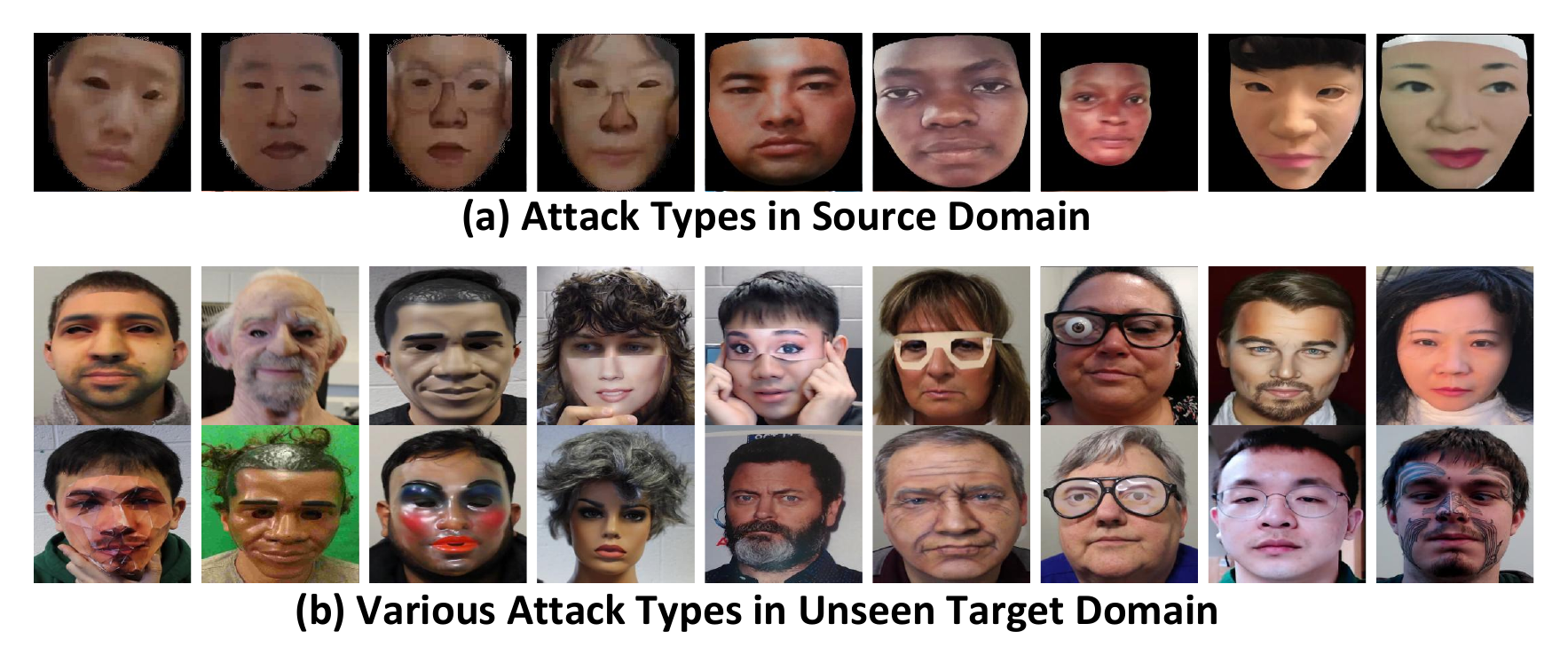} 
	\caption{Sample attacks in the source and target domains.}
	\label{fig5}
\end{figure}

\noindent \textbf{Datasets and Evaluation Protocols.}
We construct evaluation protocols using four datasets (CASIA-SURF~\citep{zhang2019dataset}, CeFa~\citep{liu2021casia}, HQ-WMCA~\citep{heusch2020deep}, SiW-Mv2~\citep{guo2022multi}) to assess model performance. Although all four datasets contain multimodal data, we utilize only the visible light modality in our experiments.
The CASIA-SURF dataset includes two types of attacks: print and cut. The CeFa dataset covers print, replay attacks, 3D print, and silicone mask attacks. The HQ-WMCA and SiW-Mv2 datasets contain ten and fourteen different attack types, respectively, many of which are not present in the CASIA-SURF or CeFa dataset. By using the CASIA-SURF and CeFa datasets as source domains and the HQ-WMCA and SiW-Mv2 datasets as target domains, face anti-spoofing scenarios characterized by significant covariate and semantic shifts are effectively constructed~\citep{chen2025mmfas,ge2024difffas}.
Examples of training and inference attack types are illustrated in \figurename~\ref{fig5}.

\noindent \textbf{Evaluation Metrics.}
We use the False Rejection Rate (FRR) and False Acceptance Rate (FAR) to evaluate the model’s detection performance on real and spoof faces, respectively. The Half Total Error Rate (HTER) is employed as a comprehensive metric to assess the model’s overall detection performance across both classes.

\noindent \textbf{Implementation details.}
Both the reinforcement fine-tuning and supervised fine-tuning methods are based on the Qwen2.5-VL-7B-Instruct model as the base model. The task prompts used in all experiments are the same, as illustrated in \figurename~\ref{fig7}. The number of samples $N$ is set to 6, the expected maximum reasoning length $L$ is set to 1200, the batch size is configured to 6, and the learning rate is set to 5e-6.

\noindent \textbf{Comparison Methods.}
For a fair comparison, we select several representative baselines, including classical traditional methods such as MS-LBP~\cite{maatta2011face}, Color texture~\cite{boulkenafet2016face}, and CNN~\cite{yang2014learn}, as well as SOTA multimodal large model-based methods such as Flip~\cite{srivatsan2023flip}, FoundPAD~\cite{ozgur2025foundpad}, and Qwen2.5-VL~\cite{bai2025qwen2}.

\subsection{Comparison with State-of-the-Art Face Anti-Spoofing Methods}
We compare the performance of our method with previous state-of-the-art face anti-spoofing approaches under cross-domain protocols, with the results presented in Table~\ref{tabcross}. Across all four protocols, it is evident that traditional methods (e.g., MS-LBP, Color texture, CNN) struggle to generalize in the presence of significant covariate and semantic shifts in the target domain. In contrast, approaches (e.g., Flip, FoundPAD) based on multimodal large models demonstrate superior performance. FoundPAD FE, which freezes the backbone and fine-tunes only the final fully connected classification layer, performs worse than both the from-scratch trained FoundPAD ViT-FS and the LoRA-tuned FoundPAD. Among the three, the LoRA-fine-tuned FoundPAD achieves the best performance in a parameter-efficient manner. Flip, which combines prompt learning with supervised parameter fine-tuning, achieves the second-best performance across three protocols, further validating the effectiveness of adapting knowledge from multimodal large models to face anti-spoofing tasks.

In terms of the HTER metric, our method achieves performance improvements of 18.48\%, 60.75\%, 18.17\%, and 28.28\% across the four protocols, respectively, compared to the second-best performing method. These results demonstrate that the proposed reinforcement fine-tuning approach effectively adapts multimodal large language models for the classification of real and spoof faces. Moreover, the fine-tuned model exhibits strong generalization capabilities in handling external multifactor variations and unseen attack types across diverse scenarios.

\begin{figure}[tb]
	\centering
	\includegraphics[width=0.46\textwidth,height=0.20\textheight ]{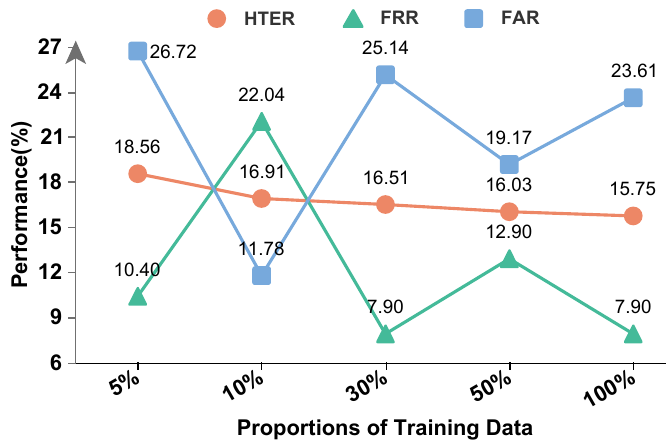} 
	\caption{Performance variation with different training data volumes under the protocol CeFa to HQ-WMCA.}
	\label{fig8}
\end{figure}
\begin{figure}[tb]
	\centering
	\includegraphics[width=0.48\textwidth]{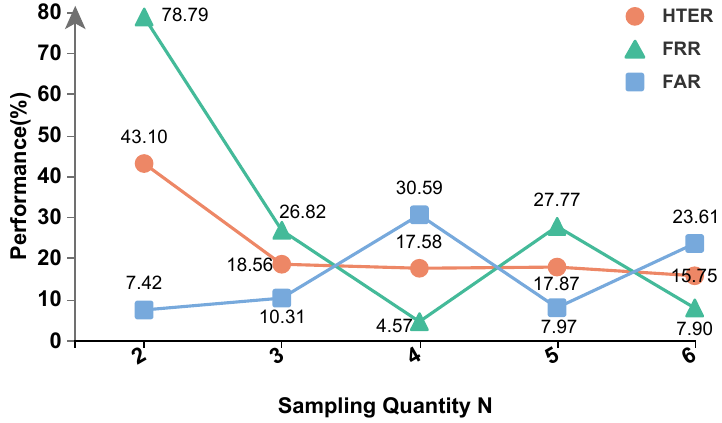} 
	\caption{Performance variation with different sampling quantities under the protocol CeFa to HQ-WMCA.}
	\label{fig9}
\end{figure}

\subsection{Comparison with Supervised Fine-Tuning}
We compare the performance of Qwen2.5-VL-7B-Instruct, its supervised fine-tuned version, and the proposed reinforcement fine-tuned version, with the results presented in Table~\ref{tabsft}. The original Qwen2.5-VL-7B-Instruct model demonstrates high accuracy in identifying real faces but performs poorly in distinguishing various types of spoof faces. After supervised fine-tuning, the model’s ability to detect spoof faces improves significantly; however, this comes at the cost of a substantial drop in its accuracy on real face detection. The discrepancy between real faces in the source and target domains primarily stems from covariate shifts caused by external scene variations. This highlights the limitation of supervised fine-tuning, which tends to memorize patterns from the source domain, making it difficult to generalize to new data distributions. In contrast, the proposed reinforcement fine-tuning method achieves robust performance in distinguishing both real and diverse spoof faces in unknown target domains. These results further confirm that reinforcement fine-tuning is more effective in enhancing the model’s generalization ability to unseen target domains.

\begin{figure*}[htb]
	\centering
	\includegraphics[width=1\textwidth,height=0.3\textheight]{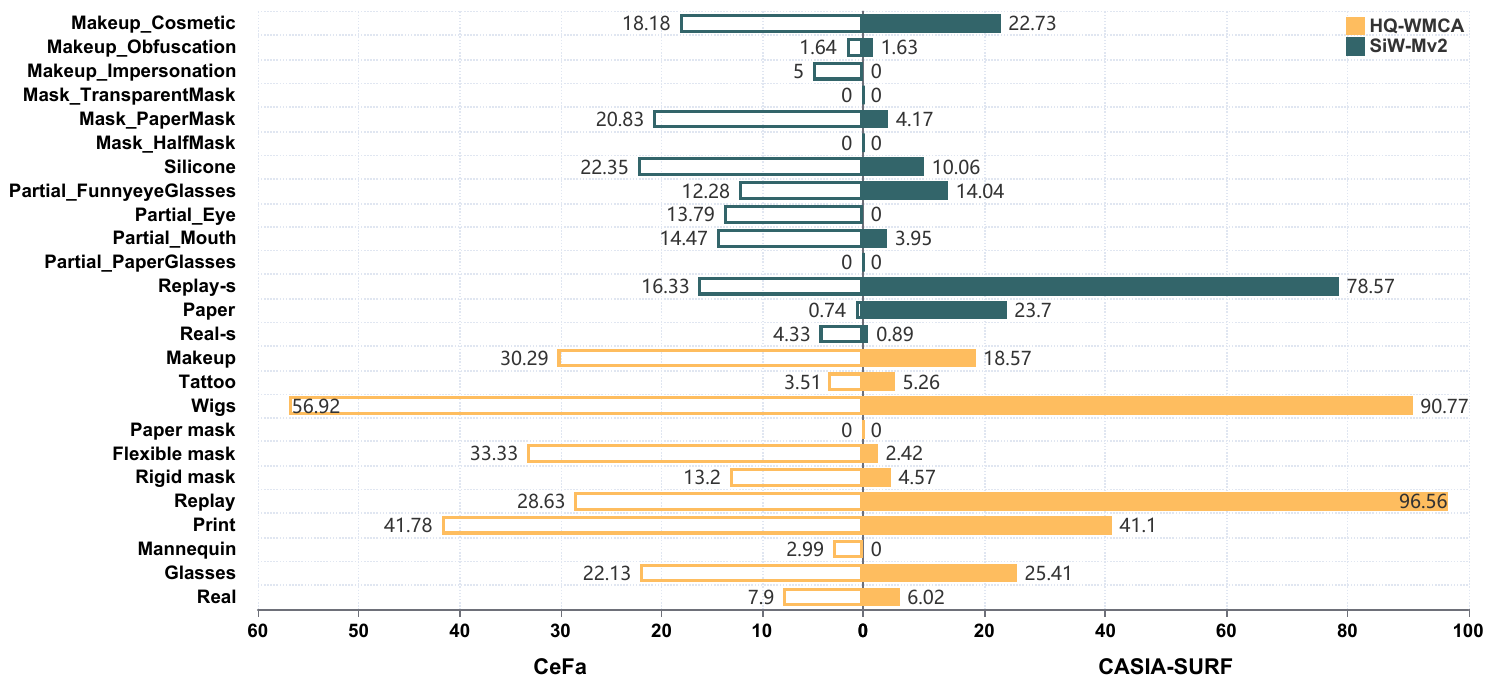} 
	\caption{Error rate(\%$\downarrow$) visualization results of various type of faces. The CeFa and CASIA-SURF datasets are the source domains and the HQ-WMCA and SiW-Mv2 datasets are the unseen target domains, respectively. }
	\label{fig1}
\end{figure*}

\begin{figure}[!htb]
	\centering
	\includegraphics[width=0.5\textwidth]{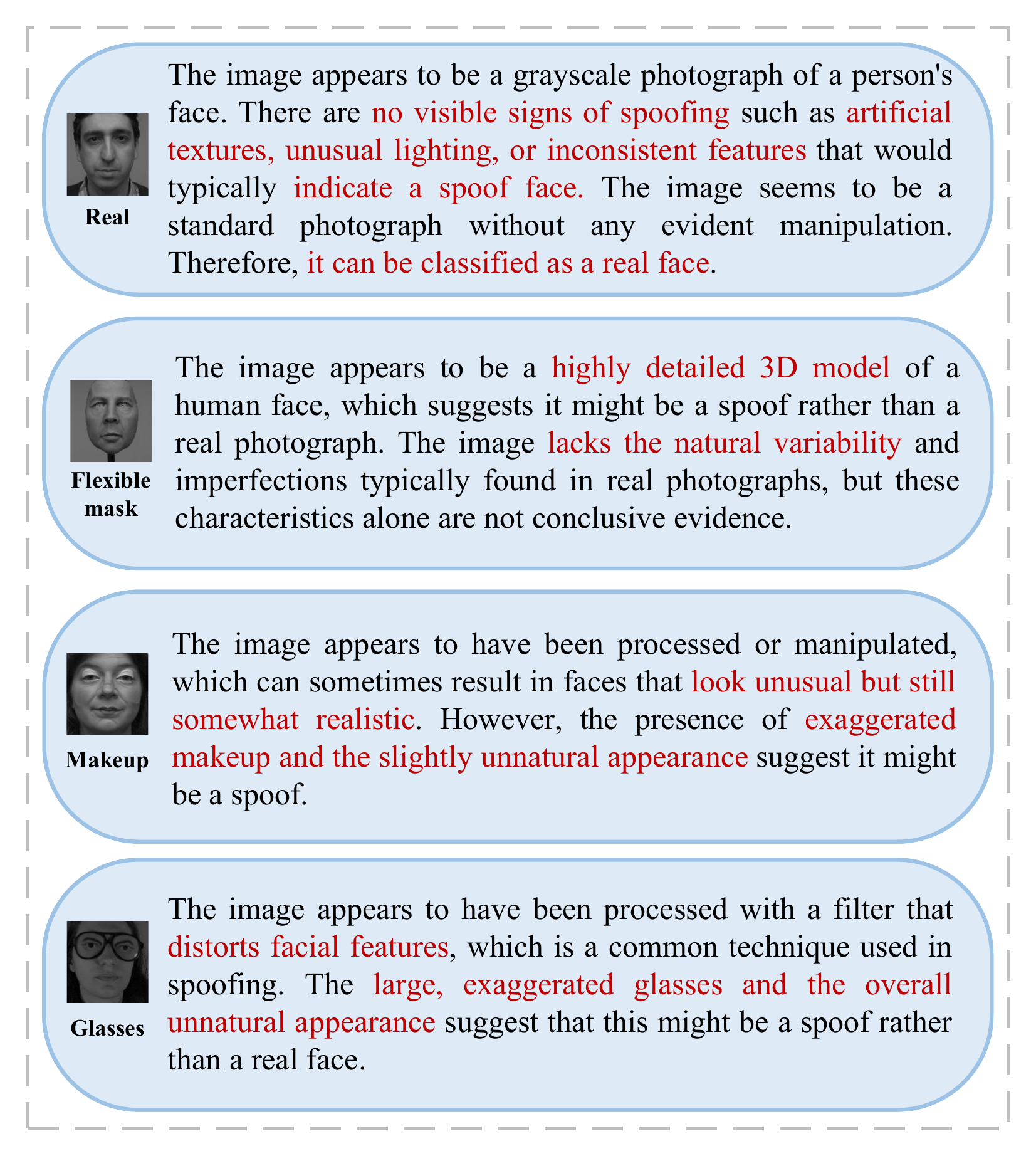} 
	\caption{Reasoning Result Visualization under the protocol CeFa to HQ-WMCA.}
	\label{fig3}
\end{figure}

\begin{table}[!tbp]
	\footnotesize
	\centering 
	\begin{tabular}{|p{2.5cm}<{\centering}|p{1.4cm}<{\centering}|p{1.4cm}<{\centering}|p{1.4cm}<{\centering}|
		}
		\hline
		& FRR(\%)$\downarrow$&FAR(\%)$\downarrow$ &HTER(\%)$\downarrow$   \\ \hline
		w/o $R_{cls}$&0.00&77.93&38.97\\ \hline
		w/o $R_{res}$&2.08&39.59&20.83\\ \hline	
		w/o $R_{format}$&32.22&5.67&18.95\\ \hline
		Ours&7.90&23.61&15.75\\
		\hline
	\end{tabular}
	\caption{Component analysis results under the protocol CeFa to HQ-WMCA.}
	\label{tabab}
\end{table}

\subsection{Ablation Study and Visualization Analysis}
\subsubsection{Component Analysis.}
We conduct ablation experiments to analyze the contribution of the three reward functions to overall performance, with the results shown in Table~\ref{tabab}. It is evident that the class consistent reward $R_{cls}$ is critical. Its removal during reinforcement fine-tuning significantly degrades the capability of the original base model. When the format reward $R_{format}$ and reasoning consistent reward $R_{res}$ are removed, the HTER metric increases by 16.89\% and 24.39\%, respectively, indicating that the format reward serves as a foundation that ensures the integrity of chain-of-thought reasoning and category decision-making. Meanwhile, the reasoning consistent reward enhances the generalization ability of models by enforcing alignment between the reasoning length and the final prediction.

\subsubsection{Impact of Training Data Volume.}
We compare the performance of models trained with varying proportions of the CeFA dataset as the source domain, with results shown in \figurename~\ref{fig8}. As reflected by the HTER metric, the generalization ability of models improves as the amount of training data increases. However, once the data volume reaches a certain threshold, the performance gains begin to plateau, indicating diminishing returns with further data expansion.

\subsubsection{Impact of Sampling Quantity.}
We compare the impact of different sampling quantities from the policy model on detection performance, with the results presented in \figurename~\ref{fig9}. As shown by the HTER metric, model performance improves with an increasing number of sampled policies. This suggests that exploring a broader range of policies facilitates the ability of models to learn more generalized solutions for face anti-spoofing tasks.

\subsubsection{Error Rate Analysis of Various Types of Faces.}
We visualize the error rates for different face types to further analyze our model’s robustness against various cross-domain attack types, as shown in \figurename~\ref{fig1} . Across the four protocols, regardless of source-target domain variations, our method demonstrates strong generalization in detecting obfuscation makeup, impersonation makeup, transparent masks, half masks, paper glasses, tattoos, paper masks, mannequins, and real faces. This indicates that reinforcement fine-tuning of multimodal large language models can yield face anti-spoofing models with strong generalization capabilities against diverse unseen attacks in cross-domain scenarios.

Notably, significant performance fluctuations are observed for replay and wig attacks when the source domain changes. The CASIA-SURF dataset does not include replay attacks, while the CeFa dataset does, highlighting that the presence of similar attack types in the fine-tuning set benefits the model’s generalization to those types. The wig attack refers to real individuals wearing wigs as a form of disguise. While the CASIA-SURF dataset lacks wig-related attacks, the CeFa dataset includes spoof faces involving wigs. This suggests that, despite the attack types not being exactly the same across domains, the reinforcement fine-tuned model 
is capable of disentangling and attributing the spoofing pattern to the presence of wigs, and effectively transferring that knowledge to handle test samples exhibiting similar spoofing characteristics.

\subsubsection{Visualization of Reasoning Results.}
Our method provides interpretable reasoning for the classification decisions between real and spoof faces. \figurename~\ref{fig3} visualizes the reasoning results for real faces and several unseen attack types from the unseen target domain. It can be observed that the model’s reasoning aligns well with common decision-making strategies in the face anti-spoofing domain. The model evaluates factors such as color, texture, lighting, distortion level, and the naturalness of facial representations. For specific types of attacks, it is also capable of accurately identifying distinguishing features, such as 3D models, exaggerated glasses, or makeup. This indicates that the reinforcement fine-tuned model has, to a certain extent, internalized the underlying logic and methodology for distinguishing between real and spoof faces.

\section{Conclusion}
In this paper, we propose a reinforcement fine-tuning-based face anti-spoofing method that harnesses the capabilities of multimodal large language models to enhance cross-domain generalization and interpretability.
Extensive experimental results demonstrate that our approach can effectively generalize to various unknown attack types in unseen target domains characterized by significant covariate and semantic shifts, while offering interpretable decision reasoning without the need for labor-intensive annotated textual explanations for training. In future work, we will explore parameter-efficient reinforcement fine-tuning strategies to further enhance the generalization capability and interpretability for cross-domain face anti-spoofing using fewer computational resources.


\bibliography{ref2}
\end{document}